\documentclass{article}
\usepackage{spconf,amsmath,graphicx}
\usepackage{multirow}
\usepackage{hyperref}

\title{Multispectral Fusion for Object Detection \\ with Cyclic Fuse-and-Refine Blocks}
\name{Heng ZHANG$^{1,3}$, Elisa FROMONT$^{1,4}$, Sébastien LEFEVRE$^{2}$, Bruno AVIGNON$^{3}$}
\address{$^{1}$Univ Rennes, IRISA $^{2}$Univ Bretagne Sud, IRISA, $^{3}$ATERMES company,$^{4}$ IUF, Inria}
\begin{document}
%
\maketitle
%


%


\begin{abstract}
Multispectral images (e.g. visible and infrared) may be particularly useful when detecting objects with the same model in different environments (e.g. day/night outdoor scenes). To effectively use the different spectra, the main technical problem resides in the information fusion process. In this paper, we propose a new halfway feature fusion method for neural networks that leverages the complementary/consistency balance existing in multispectral features by adding to the network architecture, a particular module that cyclically fuses and refines each spectral feature. We evaluate the effectiveness of our fusion method on two challenging multispectral datasets for object detection. Our results show that implementing our \textit{Cyclic Fuse-and-Refine} module in any network improves the performance on both datasets compared to other state-of-the-art multispectral object detection methods.
\end{abstract}

\section{Introduction}
\label{introduction}

\begin{figure}
\centering
\includegraphics[width=8.5cm]{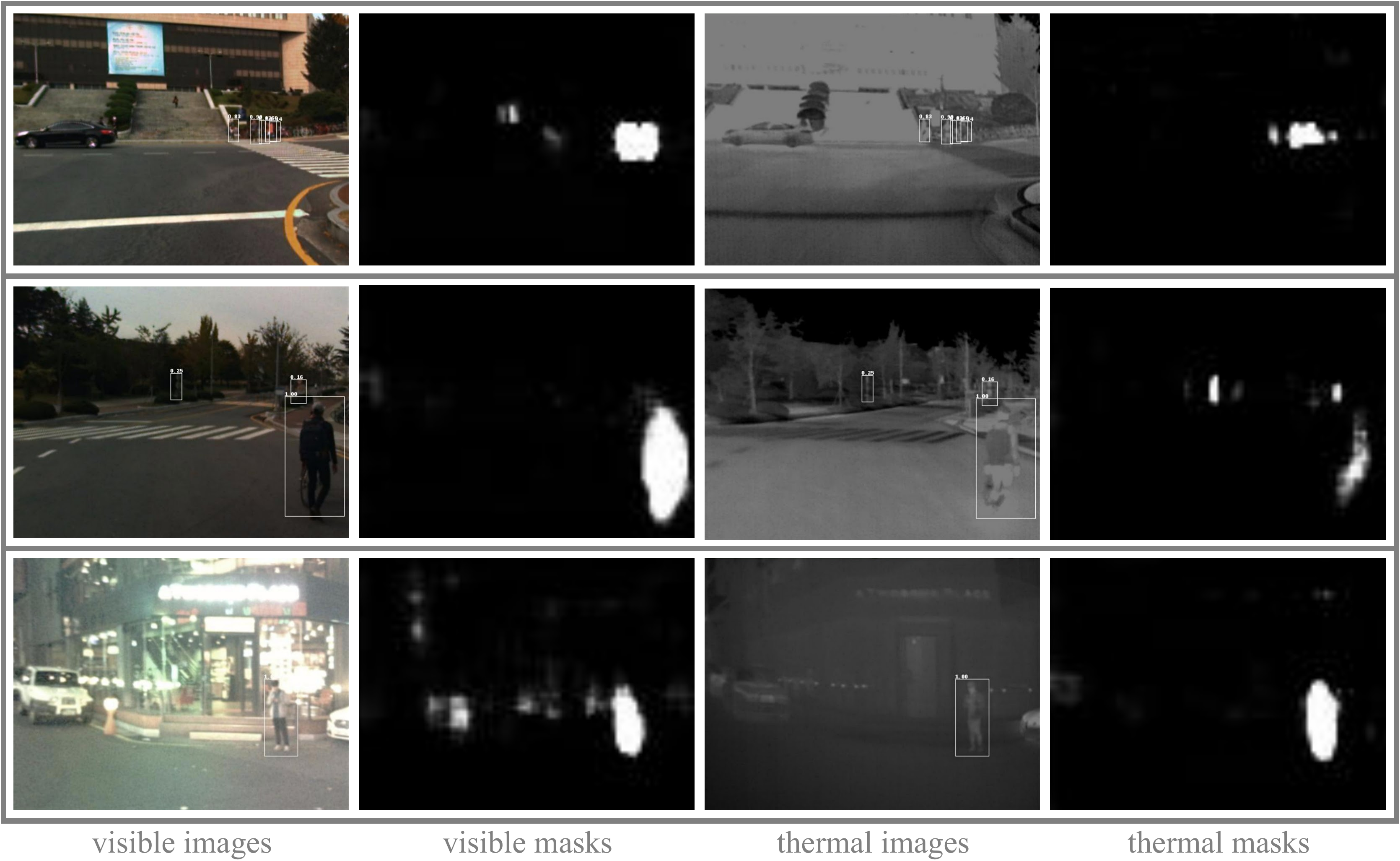}
\caption{Examples of thermal and RGB images of the same aligned scenes taken from KAIST multispectral pedestrian detection dataset \cite{KAIST} with detected bounding boxes. The segmentation masks (2nd and 4th columns) are predicted based on the (mono-)spectral features before any fusion process.}
\label{fig:verif}
\end{figure}

Visible and thermal image channels are expected to be complementary when used for object detection in the same outdoor scenes. In particular, visible images tend to provide color and texture details while thermal images are sensitive to objects' temperature, which may be very helpful at night time. However, because they provide a very different view of the same scene, the features extracted from different image spectra may be inconsistent and lead to a difficult, uncertain and error-prone fusion (Fig. \ref{fig:verif}). In this figure,  we use a Convolutional Neural Network (CNN, detailed later) to predict two segmentation masks based on the two (aligned) mono-spectral extracted features from the same image and then fuse the features to detect pedestrians in the dataset. During the training phase, the object detection and the semantic segmentation losses are jointly optimised (the segmentation ground truths are generated according to pedestrian bounding box annotations). We can observe that most pedestrians are visible either on the RGB or on the infrared segmentation masks which illustrates the \emph{complementary} of the channels. However, even though the visible-thermal image pairs are well aligned, the similarity between the two predicted segmentation masks is small, i.e., the multispectral features may be \emph{inconsistent}.

In order to augment the consistency between features of different spectra, we design a novel feature fusion approach for convolutional neural networks based on \textit{Cyclic Fuse-and-Refine} modules. Our main idea is to \emph{refine} the mono-spectral features with the \emph{fused} multispectral features multiple times consecutively in the network. Such a fusion scheme has two advantages:
1) since the fused features are generally more discriminative than the spectral ones, the refined spectral features should also be more discriminative than the original spectral features and the fuse-and-refine loop gradually improves the overall feature quality; 2) since the mono-spectral features keep being refined with the same features, their consistency progressively increases, along with the decrease of their complementary, and the consistency/complementary balance is achieved by controlling the number of loops.

\begin{figure*}[h]
\centering
\includegraphics[width=17cm]{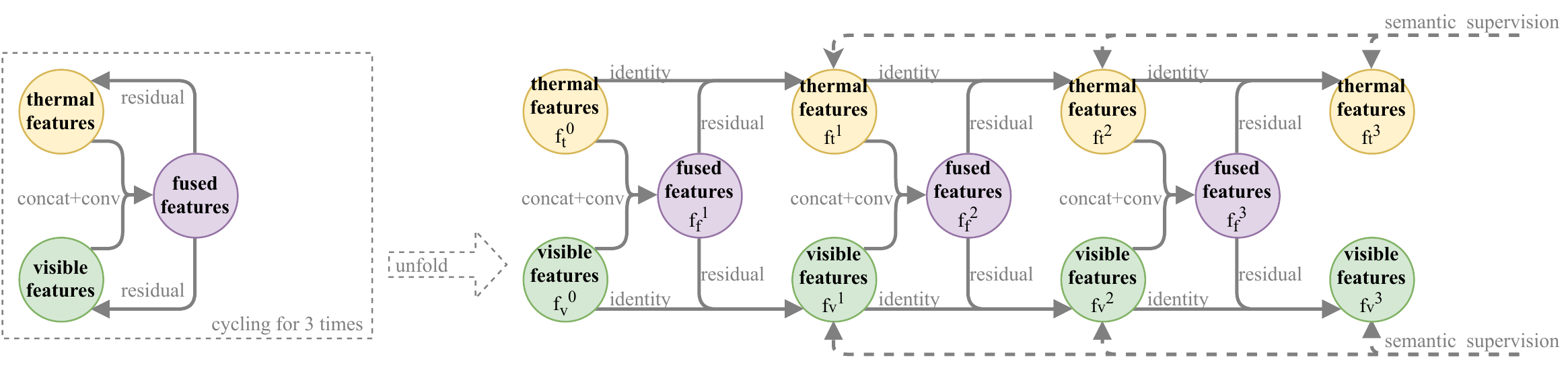}
\caption{Illustration (folded on the left part and unfolded on the right) of the proposed \textit{Cyclic Fuse-and-Refine Module} with 3 loops. Better viewed in color.}
\label{fig:fold}
\end{figure*}

We review the related works on multispectral feature fusion with CNN in Section \ref{sec:related}. We detail our novel network module named \textit{Cyclic Fuse-and-Refine}, which loops on the fuse-and-refine operations to adjust the multispectral features' complementary/consistency balance in Section \ref{sec:approach}. In Section \ref{sec:expe}, we show experiments on the well known KAIST multispectral pedestrian detection dataset \cite{KAIST} on which we obtain new state-of-the-art results, and on the less known FLIR ADAS dataset \cite{Flir} on which we set a first strong baseline.

\section{Related Work}
\label{sec:related}
Existing approaches mainly differ on the strategies (``when" and ``how") used to fuse the multispectral features.

\noindent
\textbf{When to fuse.} The first study on CNN-based multispectral pedestrian detection is made by \cite{DFCNN}, and they evaluate two fusion strategies: early and late fusions. Then \cite{TestAnno} and \cite{Illumination-aware-li} explore this further and show that a fusion of features halfway in the network, achieves better results than the early or the late fusion. Since then, the halfway fusion has become the default strategy in deep learning-based multispectral (and multimodal) works (\cite{Illumination-aware-li,Illumination-aware-guan,MSDS,CIAN,AR-CNN}). We also choose to locate our fuse-and-refine fusion module halfway in the network.

\noindent
\textbf{How to fuse.} Features extracted from each spectral channel have different physical properties and choosing how to fuse these complementary information is another central research topic. Basic fusion methods include element-wise addition/average, element-wise maximum and concatenation sometimes in addition to a $1\times1$ convolution to compress the number of channels as done e.g. in \cite{NIN}. Building on this, more advanced methods such as \cite{Illumination-aware-li} and \cite{Illumination-aware-guan} use illumination information to guide the multispectral feature fusion. \cite{GFD-SSD} apply Gated Fusion Units (GFU) \cite{GFU} to combine two SSD networks \cite{SSD} on color and thermal inputs. \cite{CIAN} propose a cross-modality interactive attention network to dynamically weight the fusion of thermal/visible features. Our strategy is  different: we suggest a cyclic fusion scheme to progressively improve the quality of the spectral features and automatically adjust the complementary/consistence balance.
\section{Proposed Approach}
\label{sec:approach}

\noindent
\textbf{Overview}. The fusion and refinement operations are the main ones of our proposed approach. They are repeated (through a cycle) multiple times to increase the consistency of the multispectral features and to decrease the complementarity of the features.
An illustration of our \textit{Cyclic Fuse-and-Refine} module with 3 loops in the cycle is presented in Fig. \ref{fig:fold}. 

\noindent
\textbf{Fuse-and-Refine}. In each loop $i$, for the fused ($_f$), visible ($_v$) and thermal ($_t$) features, the multispectral feature fusion can be formalized as $f_{f}^{i} = \mathcal{F}(\sigma(f_{t}^{i-1}, f_{v}^{i-1}))$, where $\sigma$ is a feature concatenation operation, and $\mathcal{F}$ is a $3\times3$ convolution followed by a batch normalization operation.
For simplicity and to avoid over-fitting, the operation $\mathcal{F}$ in all loops shares weights.
The fused features are then assigned as residuals of the spectral features for refinement: $f_{t}^{i} = \mathcal{H}(f_{t}^{i-1}+f_{f}^{i}), f_{v}^{i} = \mathcal{H}(f_{v}^{i-1}+f_{f}^{i})$. $\mathcal{H}$ is the activation function (e.g. ReLU).

\noindent
\textbf{Semantic supervision}. In order to prevent the vanishing gradient problem when learning the parameters of the network and to better guide the multispectral feature fusion, an auxiliary semantic segmentation task is used to bring separate supervision information for each refined spectral features. Concretely, after being refined with the fused features, the thermal and visible features go through a $1\times1$ convolution (aiming at replacing a fully-connected layer so to ensure a fully-convolutional network) to predict two pedestrian segmentation masks, one for each channel.
These predicted masks are also used to tune (or at least visualize) the number of loops in the cyclic module according to the complementary/consistency variations in the features.

\noindent
\textbf{Final fusion}. Following \cite{Deraining}, since the optimal cycling number is unknown and could be different for different image pairs, we aggregate all the refined spectral features to generate the final fused features that will be used for the object detection part of the network. The aggregation is a simple element-wise average function. Let $I$ be the number of loops, the final computation is: $\frac{1}{2 I}(\sum_{1}^{I}f_{t}^{i}+\sum_{1}^{I}f_{v}^{i})$.
\begin{figure}[]
\centering
\begin{minipage}[b]{.241\linewidth}
  \centering
  \centerline{\includegraphics[width=2.15cm]{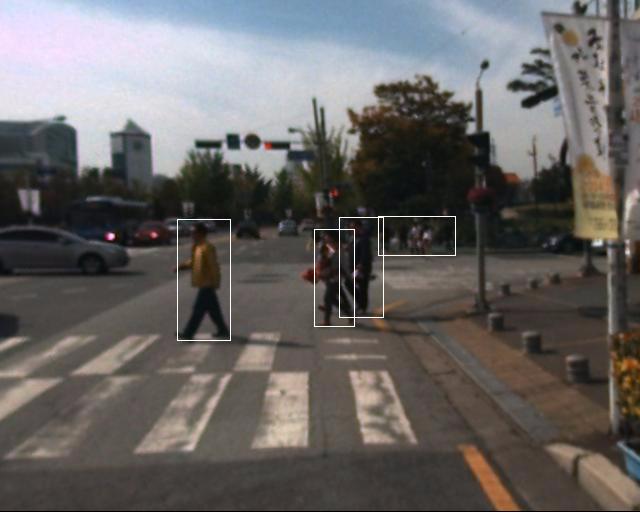}}
\end{minipage}
\begin{minipage}[b]{.241\linewidth}
  \centering
  \centerline{\includegraphics[width=2.15cm]{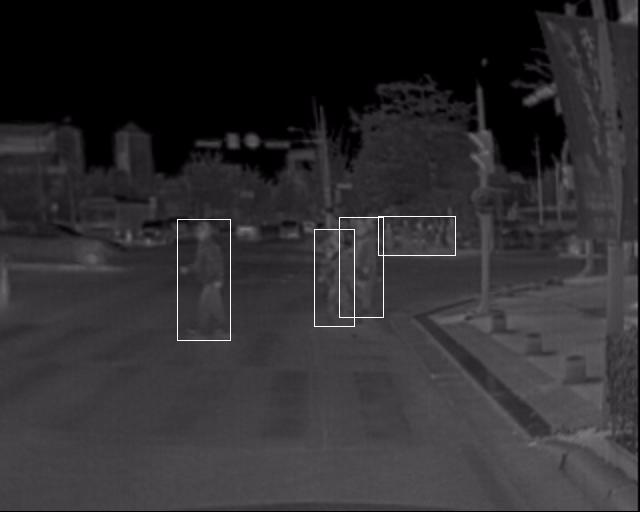}}
\end{minipage}
\begin{minipage}[b]{.241\linewidth}
  \centering
  \centerline{\includegraphics[width=2.15cm]{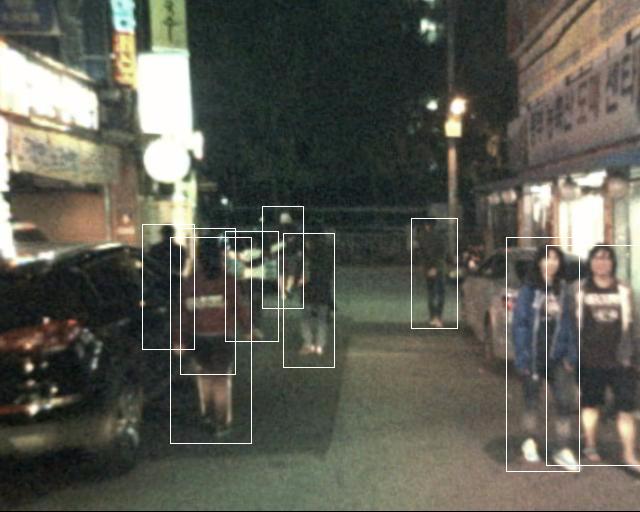}}
\end{minipage}
\begin{minipage}[b]{.241\linewidth}
  \centering
  \centerline{\includegraphics[width=2.15cm]{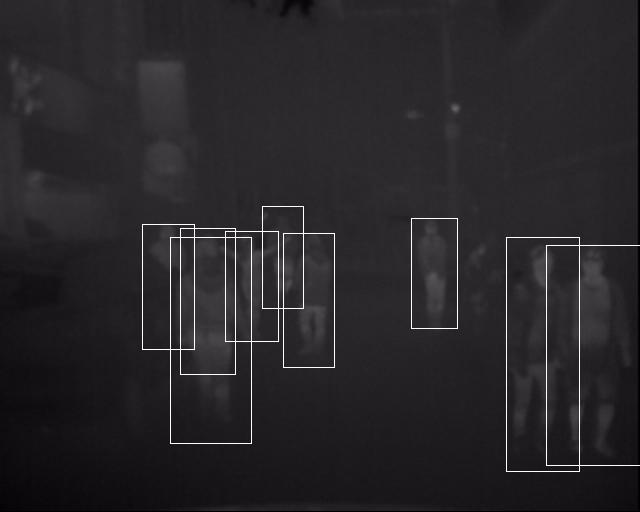}}
\end{minipage}
\begin{minipage}[b]{.241\linewidth}
  \centering
  \centerline{\includegraphics[width=2.15cm]{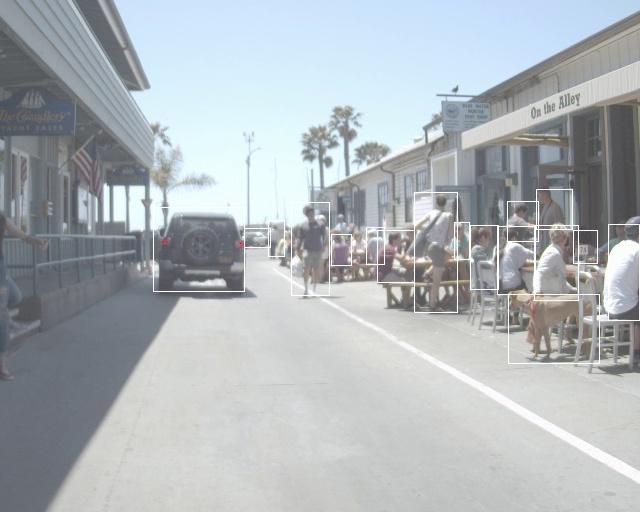}}
\end{minipage}
\begin{minipage}[b]{.241\linewidth}
  \centering
  \centerline{\includegraphics[width=2.15cm]{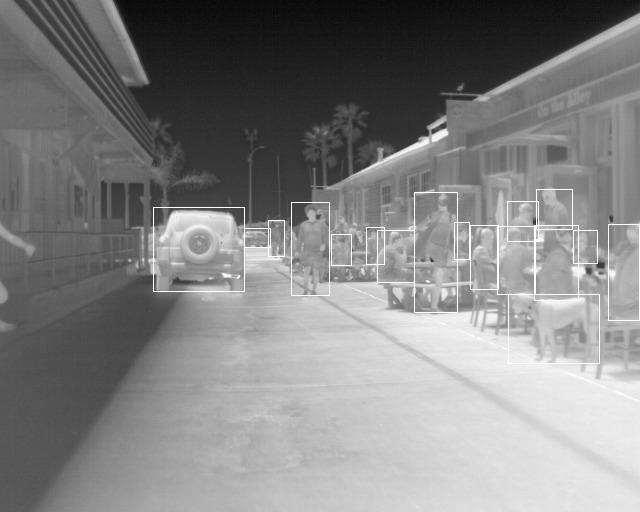}}
\end{minipage}
\begin{minipage}[b]{.241\linewidth}
  \centering
  \centerline{\includegraphics[width=2.15cm]{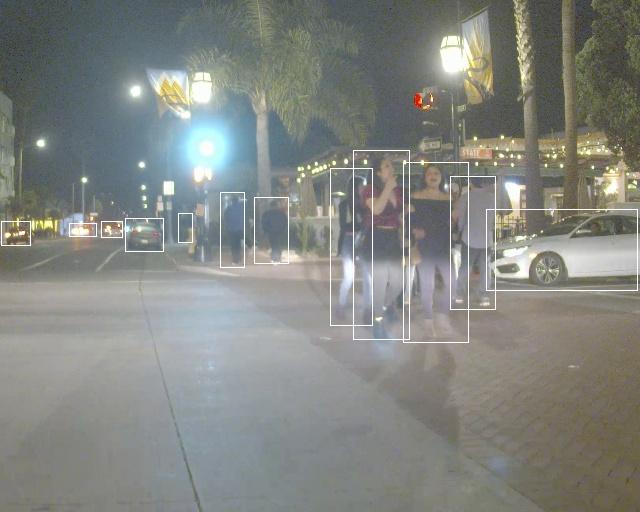}}
\end{minipage}
\begin{minipage}[b]{.241\linewidth}
  \centering
  \centerline{\includegraphics[width=2.15cm]{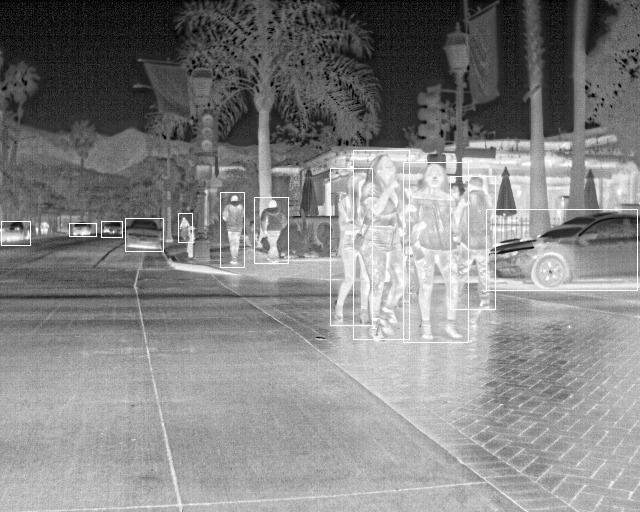}}
\end{minipage}
\begin{minipage}[b]{.241\linewidth}
  \centering
  \centerline{\includegraphics[width=2.15cm]{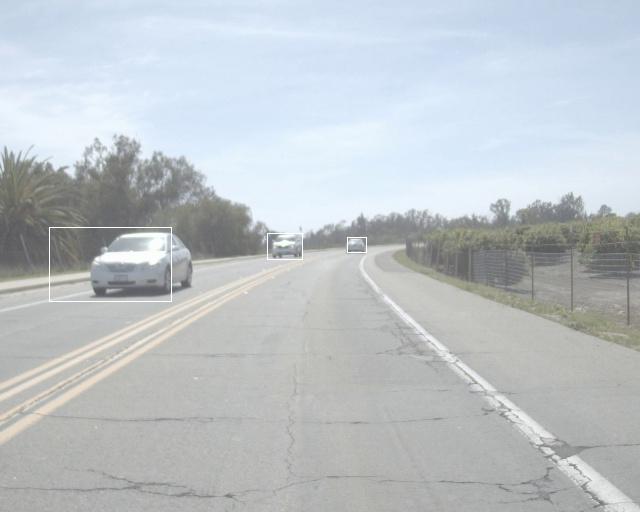}}
\end{minipage}
\begin{minipage}[b]{.241\linewidth}
  \centering
  \centerline{\includegraphics[width=2.15cm]{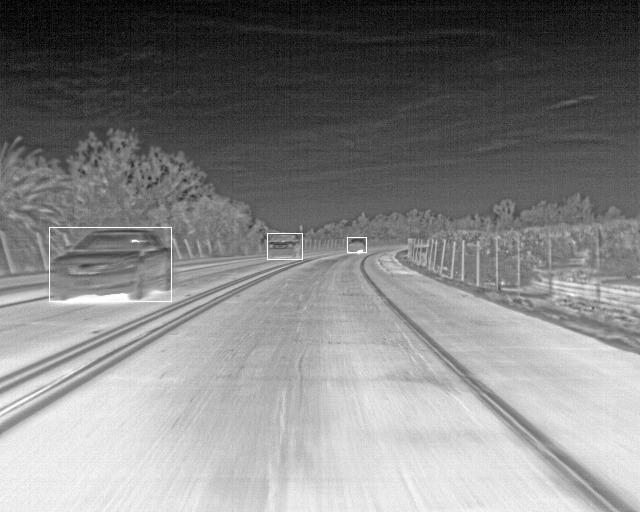}}
\end{minipage}
\begin{minipage}[b]{.241\linewidth}
  \centering
  \centerline{\includegraphics[width=2.15cm]{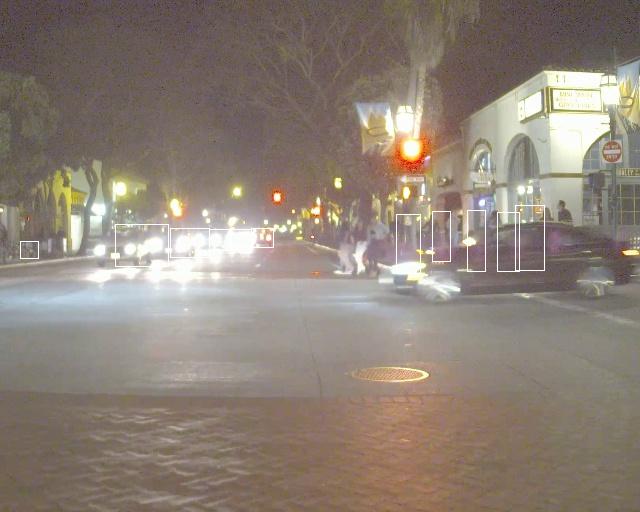}}
\end{minipage}
\begin{minipage}[b]{.241\linewidth}
  \centering
  \centerline{\includegraphics[width=2.15cm]{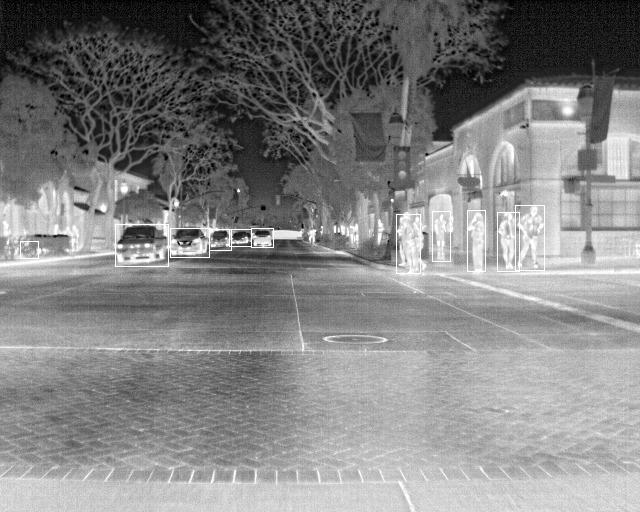}}
\end{minipage}
\caption{Examples of visible/thermal image pairs with their ground truth from the KAIST dataset (in the first line) and from the FLIR dataset in the second and third lines (the ground truth annotations are given according to the thermal images). The third line gives an example of misaligned pairs in the FLIR dataset. 
Better viewed in color and zoomed in.}
\label{fig:example}
\end{figure}

\section{Experiments}
\label{sec:expe}

We evaluate the proposed \textit{Cyclic Fuse-and-Refine Module} on KAIST Multispectral Pedestrian Detection \cite{KAIST} and FLIR ADAS dataset \cite{Flir},
and compare our results with the state-of-the-art multispectral methods. Examples of  image pairs with their ground truth bounding boxes are shown in Fig.~\ref{fig:example}.

\subsection{Datasets}
 
\noindent
\textbf{KAIST.} We use the processed version of this multispectral pedestrian detection dataset which contains 7,601 color-thermal image pairs for training and 2,252 pairs for testing. We kept the bounding boxes annotated as ``person", ``person?" or ``people" as positive pedestrian examples. \cite{MSDS} proposed a ``sanitized" version of the training annotations which eliminated some of the annotation errors from the original training annotations.
According to \cite{TestAnno}, inaccurate annotations in the test set leads to unfair comparisons, so we only use their ``sanitized" testing annotations for our evaluation, with the usual ``Miss Rate'' performance metric under reasonable setting, i.e., a test subset containing not/partially occluded pedestrians which are larger than 55 pixels.

\noindent
\textbf{FLIR.} This recently released multispectral (multi-)object detection dataset contains around 10k manually-annotated thermal images with their corresponding reference visible images, collected during daytime and nighttime. We only kept the 3 more frequent classes which are ``bicycle", ``car" and ``person". We manually removed the misaligned visible-thermal image pairs and ended with 4,129 well-aligned image pairs for training and 1,013 image pairs for test \footnote{This new aligned dataset can be downloaded here: \url{http://shorturl.at/ahAY4}}. Some examples of the well-aligned and misaligned visible-thermal image pairs are shown in Figure \ref{fig:example}.

\subsection{Training details}

\noindent
\textbf{Network architecture.} We implemented our \textit{Cyclic Fuse-and-Refine} module on the single stage object detector FSSD \cite{FSSD}, which is an improved version of the well known SSD object detector \cite{SSD}. Note that our proposed module is independent from the chosen network architecture. Following \cite{TestAnno} and \cite{Illumination-aware-li}, the mono-spectral features are extracted independently through a VGG16 \cite{VGG} network, and fused after the conv4\_3 layer (halfway through the network). Our baseline architecture uses the element-wise average for the multispectral feature fusion and we integrate and evaluate the proposed module with different number of loops.

\noindent
\textbf{Data augmentation.} As implemented in SSD \cite{SSD} and FSSD \cite{FSSD}, a few data augmentation methods are applied, such as image random cropping, padding, flipping and distorting for both visible and thermal images. 

\noindent
\textbf{Anchor designing.} Following \cite{fourone}, the anchor designing strategy is adapted for the pedestrian detection for KAIST dataset: we fix the aspect ratio of each anchor box to $0.41$ and we only keep three detection layers with scales $32$ and $32\sqrt{2}$, $64$ and $64\sqrt{2}$, $128$ and $128\sqrt{2}$ from fine to coarse respectively. For FLIR, we use the same scale settings but we augment the aspect ratio setting to $\{1,2,\frac{1}{2}\}$.

\noindent
\textbf{Loss functions.} To improve object detection, SDS RCNN \cite{SDS} and MSDS RCNN \cite{MSDS} use an additional task, semantic segmentation, and jointly optimize the loss for the segmentation and detection tasks while training the network. To fairly compare our work to these competitors, we also use this auxiliary loss to supervise the training of the proposed module.

\subsection{Comparison with state-of-the-art methods}

\textbf{On KAIST.} We compare the experimental results of our approach with state-of-the-art methods in Table \ref{tab:sota}.
For these experiments, we make 3 loops in the Fuse-and-Refine cycle. Depending on what was done in the literature and to allow a fair comparison, we report our detection accuracy with sanitized and original training annotations respectively. 
All the deep learning-based methods \cite{TestAnno,RPNM,Illumination-aware-li,Illumination-aware-guan,MSDS} use the same input image resolution ($640\times512$) and the same backbone network (VGG16). The results show that our proposed method allows us to obtain better detection results than all its competitors for both the sanitized and original training annotations.
Note that the computational overhead from \textit{CFR} is quite small. During inference, each cycle only add $\sim$0.4ms of inference time. 

\noindent
\textbf{On FLIR.} Because of the misalignment problems in the dataset, there is, to our knowledge, no paper which uses the FLIR dataset \cite{Flir} for multispectral object detection. We use our sanitized version of the dataset and compare the mAP percentage of two different models: a baseline model which uses the traditional halfway fusion architecture (with the VGG backbone) and the same model with our proposed module. Again, we can see in Table \ref{tab:flir} that our method provides important mAP gains for all the considered object categories.

\begin{figure}[]
\centering
\includegraphics[width=8.5cm]{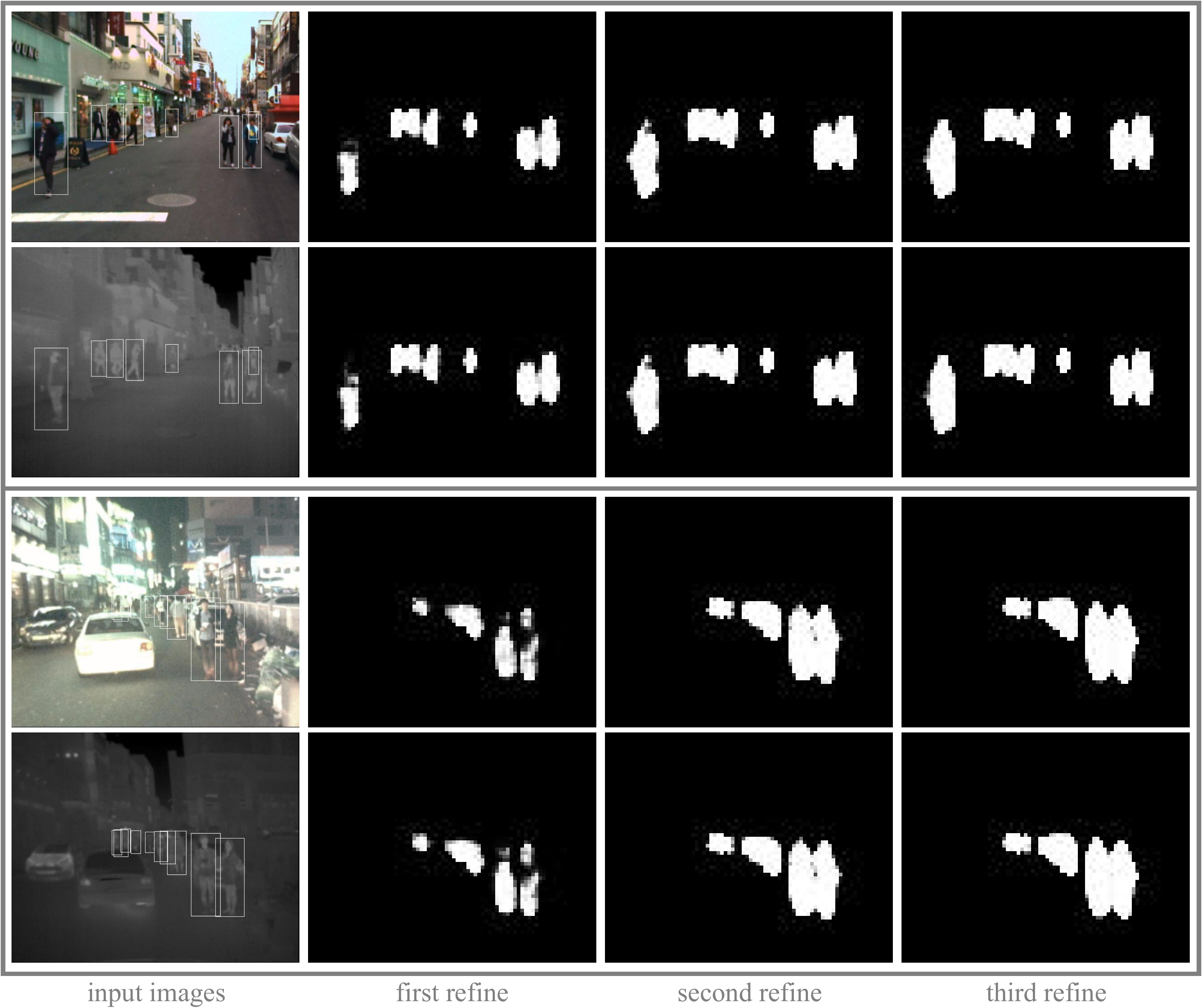}
\caption{Examples of pedestrian segmentation masks predicted on 2 visible/thermal image pairs (one taken at day time, one taken at night) of the KAIST dataset after a different number of loops (1-3) in the fuse-and-refine cycle.}
\label{fig:refine3}
\end{figure}

\subsection{Ablation study}
\label{abstudy}
We study in details (on the KAIST dataset with the sanitized training annotations and the reasonable test set) the effectiveness of the proposed fusion module and the relationship between the number of loops in the fuse-and-refine cycle and the multispectral feature complementary/consistency balance.
The experimental results are summarised in Table \ref{tab:ablation}. We provide the Miss Rate and DICE scores \cite{DICE} between the pedestrian masks predicted by each version of the refined thermal/visible features. These DICE scores are used as an indicator of similarity between the spectral features.
From the table we observe successive accuracy gains from the baseline (no loop) to 3 loops, and a decrease after 4 loops; meanwhile the value of DICE scores continue to increase along with the number of loops.
We then visualize, on two sample image pairs, the pedestrian masks predicted by visible/thermal features after each refinement in Figure \ref{fig:refine3}. The first column corresponds to input images marked with the detected pedestrians; The second, third and fourth columns correspond to segmentation masks predicted after 1 to 3 loops. The first and third lines (resp. second and fourth) are for visible (res. thermal) images and their corresponding segmentation masks. It can be observed that the quality and similarity of the masks gradually increase with the number of loops. With the increase of similarity between the spectral features, their consistency increases and their complementarity decreases. As mentioned in Section \ref{introduction}, the lack of consistency between the multispectral features is harmful; on the contrary, too much consistency leads to sharp emerge/plunge in the feature values, and makes the fusion meaningless. That explains why the Miss Rate starts to decrease after 4 loops. In practice the number of loops should be tuned for any dataset but we believe that very few values should be tried (between 2 and 5). 

\begin{table}[h!]
\centering
\begin{tabular}{llll}
\hline
\multicolumn{1}{c}{\multirow{2}{*}{Methods}} & \multicolumn{3}{c}{Miss Rate (lower, better)}\\ \cline{2-4}
\multicolumn{1}{c}{} & \multicolumn{1}{c}{R-All} & R-Day & \multicolumn{1}{c}{R-Night} \\ \hline
\multicolumn{4}{l}{\textit{Training with sanitized annotations:}}  \\
MSDS-RCNN \cite{MSDS} & 7.49\% & 8.09\% & 5.92\% \\
CFR\_3 & \textbf{6.13\%} & \textbf{7.68\%} & \textbf{3.19\%} \\ \hline
\multicolumn{4}{l}{\textit{Training with original annotations:}} \\
ACF+T+THOG \cite{KAIST} & 47.24\% & 42.44\% & 56.17\% \\
Halfway Fusion \cite{TestAnno} & 26.15\% & 24.85\% & 27.59\% \\
Fusion RPN+BF \cite{RPNM} & 16.53\% & 16.39\% & 18.16\% \\
IAF R-CNN \cite{Illumination-aware-li} & 16.22\% & 13.94\% & 18.28\% \\
IATDNN+IASS \cite{Illumination-aware-guan} & 15.78\% & 15.08\% & 17.22\% \\
MSDS-RCNN \cite{MSDS} & 11.63\% & 10.60\% & 13.73\% \\
CFR\_3 & \textbf{10.05}\% & \textbf{9.72\%} & \textbf{10.80\%} \\ \hline
\end{tabular}
\caption{
Detection accuracy comparisons in terms of Miss Rate percentage on KAIST Dataset \cite{KAIST}. Our competitors' results are taken from \cite{Illumination-aware-li} and \cite{MSDS}.
}
\label{tab:sota}
\end{table}

\begin{table}[h!]
\centering
\begin{tabular}{lllll}
\hline
\multicolumn{1}{c}{Methods} & \multicolumn{1}{c}{mAP} & \multicolumn{1}{c}{Bicycle} & Car & \multicolumn{1}{c}{Person} \\ \hline
Baseline & 71.17\% & 56.39\% & 83.90\% & 73.28\% \\
CFR\_3 & \textbf{72.39\%} & \textbf{57.77\%} & \textbf{84.91\%} & \textbf{74.49\%} \\ \hline
\end{tabular}
\caption{
mAP results for two CNN object detection architectures which use (or not) our Cyclic Fuse-and-Refine (CFR) blocks on FLIR dataset \cite{Flir}.
}
\label{tab:flir}
\end{table}

\begin{table}[h!]
\centering
\begin{tabular}{lll}
\hline
Methods & Miss Rate & DICE Scores \\ \hline
Baseline & 7.68\% & - \\
CFR\_1 & 6.90\% & \{64.53\%\} \\
CFR\_2 & 6.40\% & \{78.89\%, 89.70\%\} \\
CFR\_3 & 6.13\% & \{74.60\%, 90.60\%, 94.17\%\} \\
CFR\_4 & 7.09\% & \{58.25\%, 85.91\%, 92.9\%, 96.11\%\} \\ \hline
\end{tabular}
\caption{Miss rates versus DICE scores w.r.t. different numbers of Fuse-and-Refine loops. Each experiment is repeated five times and we report the average performance.}
\label{tab:ablation}
\end{table}

\section{Conclusion}
\label{sec:conclusion}
This paper proposes a novel \textit{cycle fuse-and-refine} module to improve the multispectral feature fusion while taking into account the complementary/consistency balance of the features. Experiments on KAIST \cite{KAIST} and FLIR \cite{Flir} datasets show that integrating the proposed fusion module to a ``vanilla" multispectral pedestrian detector leads to substantial accuracy improvements.
Several visible/thermal image pairs have a misalignment problem in FLIR dataset. This problem could be more serious in real world applications due to calibration errors or temporal shifts. A Region Feature Alignment (RFA) module \cite{Disparity} tackled such a cross-modality disparity problem in a supervised manner and in a two-stage object detection setting. In the future, we would like to explore a more general solution to this problem with a similar cyclic-align scheme.
\clearpage

\bibliographystyle{IEEEbib.bst}
\bibliography{refs}


\end{document}